# Information And Control: Insights from within the brain

*Invited keynote delivered at **The Seventeenth International Multi-Conference on Computing in the Global Information Technology**, ICCGI 2022. May 22-26 2022 ,Venice, Italy.*

Birgitta Dresp-Langley
Centre National de la Recherche Scientifique (CNRS)
UMR 7357 ICube Lab Strasbourg University
Strasbourg, France
e-mail: birgitta.dresp@unistra.fr

*Abstract*—**The neural networks of the brain are capable of learning statistical input regularities on the basis of synaptic learning, functional integration into increasingly larger, interconnected neural assemblies, and self-organization. This self-organizing ability has implications for biologically inspired control structures in robotics. On the basis of signal input from vision, sound, smell, touch and proprioception, multisensory representations for action are generated on the basis of physically specified input from the environment. The somatosensory cortex is a brain hub that delivers a choice example of integration for multifunctional representation. All sensory information is in a first instance topologically represented in the brain, and thereafter integrated in somatosensory neural networks for multimodal and multifunctional control of complex behaviors. Multisensory input triggers interactions between visual, auditory, tactile, olfactive, and proprioceptive mechanisms, which cooperate or compete during brain learning, and contribute to the formation of integrated representations for action [3], reflection, and communication between the brain and the outside world. Interactions between the brain and the world enable complex actions and produce further learning and cognition. Multimodal integration is fueled by stimulation and direct action, and enables coherent brain representation of an intrinsically ambiguous physical environment.**

BACKGROUND

In physical theory theory, information is aimed at resolving uncertainty in complex systems, like the Turing machine, which inspired the early approaches The premises therein, presumably for lack of insight from neuroscience, Artificial Intelligence (AI) and quantum computers as we know them today, neither specify the nature of information, nor what we have to understand by complexity or uncertainty. This keynote summarizes insights from biological synapses in the brain, contemporary neuroscience, and neurophilosophy to illustrate why and how the concepts of information, complexity and uncertainty are reflected in "natural" (synapses, cells, brains, minds) and "artificial" (machines) systems. The context dependency of information processing by brain networks is brought forward. It allows us to understand "information" in terms of a functional link between the brain and the world, which has implications for AI, at different levels of functional complexity and/or autonomy. The synaptic activity levels of functionally connected neurons in the brain determine the activity levels of the whole functional network. Network connectivity is established by synaptic learning and self-organizes on the basis of life-long interaction with the world (Figure 1). Over the recent decade a growing interest in multisensory integration and action control has been witnessed, especially in connection with the idea of a statistically optimized integration of multiple sensory sources. The human information processing system, i.e. the brain, is to adjust moment-by-moment the relative contribution of each sense's estimate to a multisensory task. The sense's contribution depends on its variance, so that the total variance of the multisensory estimate is lower than that for each sense alone. Accordingly, the validity of a statistically optimized multisensory integration has been demonstrated by extensive empirical research, also in applied settings with as tool-use. It will be helpful to delve deeper into the multisensory information processing mechanisms and their neural correlates, asking about the range and constraints of these mechanisms, about its localization and involved networks. The contributions to the present research topic range from how information from different senses and action control are linked and modulated by specific object affordances, sometimes task-irrelevant information, temporal and spatial coupling within and between senses developing in childhood development for multisensory brain control. Correspondences play an important role. Integration does not take place when vision and touch are spatially separated. Cognitive approaches on action effect control assume that information from different senses is still coded and represented within the same cognitive domain, when the information is relevant to one and the same action. This keynote will not address the corresponding issue of modality-specific action control.

CONCEPTUAL DEVELOPMENT

Brain integration relies on synaptic learning. During this process, the synaptic weights of neural connections are either reinforced or suppressed depending on statistical properties of the input from the environment. Synaptic learning enables the formation of increasingly complex connectivity between functionally specific neural networks (Figure 1).

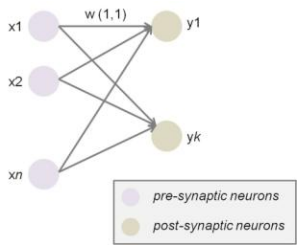

Fig.1 Synaptic weight modification during learning

It is further assumed that functionally specific organization from synapses to networks forms the brain basis of, and further promotes, self-organized learning, long-range connectivity, and multifunctional integration (Figure 2).

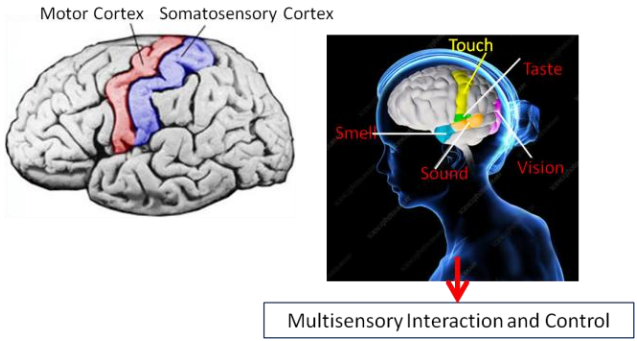

Fig. 2 The somatosensory cortex as a brain hub for multimodal integration

The functional organization of neural networks and their connectivity is largely context dependent. In the case of somatosensation, the same receptor regions of the body will activate different neural network connections under identical local stimulation when the context of the stimulation changes (Figure 3).

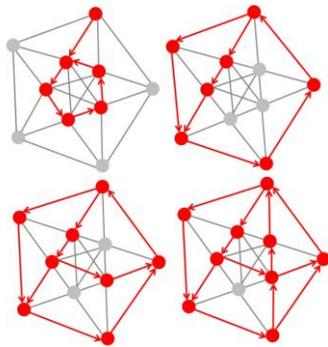

Fig. 3. Different neural network representations are generated in response to one and the same target stimulus when the context of the stimulation changes

## MATERIALS AND METHODS

Context dependent functional brain organization has measurable behavioral consequences that can be brought to the fore systematically, i.e. psychophysically, in specifically tailored experiments on human individuals performing tasks that are sensitive to multiple sensory input from the environment. Here, I will discuss the example of a motor task executed by healthy participants under different sensorial conditions. The sensory context effect on the hand grip forces produced by eight young men is measured using wireless grip force sensor technology and computer controlled data collection and analysis. 8 healthy men aged 20-30 years were tested individually performing a power grip task in repeated sessions of 10 seconds each. The grip force task was performed to either soft or hard music, with both eyes open or blindfolded (Figure 3). The men were holding two handles of equal weight, wearing wireless grip force sensor gloves for hand grip force measurement. After careful calibration, the data are sent from the gloves to the computer via blue-tooth enabled Arduino.

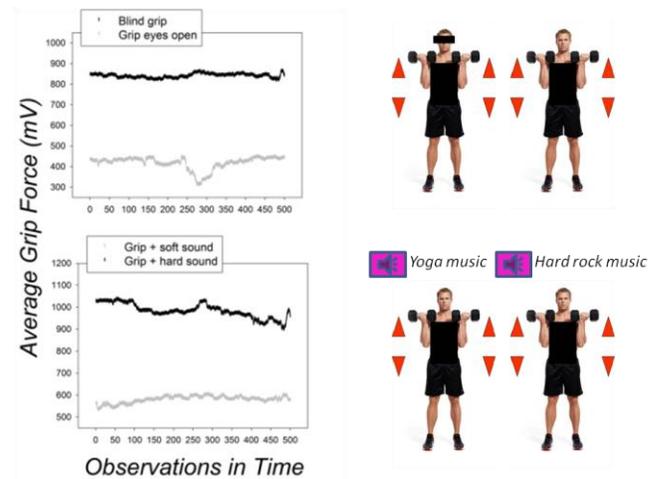

Fig. 3  Grip forces from specific sensor locations (graphs on left) measured under different sound and vision conditions (illustrations on right).

## DISCUSSION

The integration of multisensory information is an essential mechanism in perception and action control. When the grip force task is executed to loud hard rock music, the grip forces generated in certain, functionally specific sensors are considerably stronger, which is reflected here in thei example by higher measurement values in millivolt (Figure 3, bottom left). When the grip force task is executed blindfolded (no direct visual brain activation), the grip forces generated in other functionally specific sensors are considerably stronger, which also translates into higher measurement values in millivolt (Figure 3, top left). Research in multisensory integration is concerned with how contextual modulation through multiple sources of stimulation produces representations and controls behavior.

The information from the different sensory modalities, such as the senses of vision, hearing, smell, taste, touch, and proprioception, are integrated to a coherent representations of objects and actions related to these objects. The more information our brain has integrated about an object and its properties in a variety of contexts, the more adapted will be our behavior (actions on the object etc.). The self-organized functional growth of brain networks does not require adding structural components (cells, neurons). The brain as a system can grow and develop new, dynamic functionalities without the need for adding further structural complexity. As the external environment of the system changes, functional components or cells directly interacting with the environment will adapt their state(s) in a self-reinforcing process to maintain their fitness within the system. This adaptive fit will propagate further inwards, until the whole functional structure is fully adapted to the new situation. Thus, a dynamically growing self-organizing system constantly re-organizes by mutually balancing internal and external pressures for change while trying to maintain its general functional organization, and to counteract any loss thereof. Functional self-preservation is, indeed, a self-organizing system's main purpose, and each component or cell is adaptively tuned to perform towards this goal. A self-organized system is stable, largely scale-invariant, and robust against adverse conditions. At the same time, it is highly dynamic.

CONCLUSIONS

Combination of information from the different senses is central for action control. The context dependency of information processing by brain networks was brought forward here under the light of a specific example of somatosensory processing and brain integration for the control of a complex behavior in variable contexts. The illustration leads us to understand information in terms of a functional link between the brain and the world, which has implications for AI, at different levels of functional complexity and/or autonomy. Self-organization is a major functional principle that allows us to understand how the human brain works, and to establish clear functional link the "mental" and the "physical". As a fundamental conceptual support in computational thinking permits the design of a variety of modular systems, and helps achieve system stability and reliability while reducing complexity. It enables thinking in terms of self-reinforced systemic learning where the activity of connections directly determines their performance, allowing for the conceptual design of a variety of adaptive systems that are able to learn and grow independently without the need for adding non-necessary structural complexity. The right balance of structural and functional complexity conveys functional plasticity, allowing the system to stabilize but, at the same time, remain functionally dynamic and able to learn new data. Activity-dependent functional systemic growth in minimalist sized network structures is another strong advantage of self-organizing system, as it reduces structural complexity to a minimum and promotes dimensionality reduction, which is a fundamentally important quality to the development of reliable Artificial Intelligence. Bigger neural networks akin to those currently used for deep learning do not necessarily learn better or perform better. Self-organization is therefore the key to designing networks that will learn increasingly larger amounts of data increasingly faster as they learn, consolidate what has been learnt, and generate output that is predictive instead of being just accurate. In this respect, the principle of self-organizing will help design Artificial Intelligence that is not only reliable, but also meets the principle of scientific parsimony, where complexity is minimized and functionality optimized.